# Hybrid LLM/Rule-based Approaches to Business Insights Generation from Structured Data

**Aliaksei Vertsel and Mikhail Rumiantsau**
Narrative BI (team@narrative.bi)

## Abstract

In the field of business data analysis, the ability to extract actionable insights from vast and varied datasets is essential for informed decision-making and maintaining a competitive edge. Traditional rule-based systems, while reliable, often fall short when faced with the complexity and dynamism of modern business data. Conversely, Artificial Intelligence (AI) models, particularly Large Language Models (LLMs), offer significant potential in pattern recognition and predictive analytics but can lack the precision necessary for specific business applications. This paper explores the efficacy of hybrid approaches that integrate the robustness of rule-based systems with the adaptive power of LLMs in generating actionable business insights.

## 1. Introduction

As organizations grapple with increasingly complex and diverse data sets, the demand for advanced techniques that can extract valuable insights has grown exponentially. Traditional rule-based systems have often struggled to keep up with the intricacies of modern business data, while stand-alone AI Models, although powerful, may still have limitations in certain scenarios.

In response to these challenges, the concept of hybrid approaches has emerged as a compelling solution. By combining the strengths of rule-based systems and AI models, hybrid approaches offer the potential to enhance the process of data extraction and uncover meaningful insights from diverse data sources. In this paper, we explore the use of LLM-powered and rule-based systems to address the complexities of data extraction in the field of business intelligence.

The following sections will investigate the details of this hybrid approach, assessing its effectiveness in navigating the complexities of business data and extracting actionable insights.

## 2. The Hybrid LLM-Powered and Rule-Based Approach

The hybrid approach combines the strengths of interpretable AI techniques, such as LIME, rule-based systems, and supervised document classification [1], to create a powerful framework for extracting actionable insights from business data. LLM, or Large Language Model, plays a vital role in the extraction process by integrating with rule-based systems to enhance the understanding and generation of natural language-generated insights. By utilizing LLM's ability to model linguistic characteristics and generate coherent responses, the hybrid approach can uncover personalized and nuanced user interests, needs, and goals from user journeys and user activities on the platform [2].

LLM, as an interpretable AI method, can explain the individual predictions of black-box ML models and provide insight into the underlying decision-making process. This can greatly enhance the transparency and trustworthiness of the data extraction process, as stakeholders can easily understand and validate the generated insights. Furthermore, the hybrid LLM-powered and rule-based approach utilizes the logic learning machine technique for supervised data mining.

### Considerations

When implementing a hybrid LLM-powered and rule-based approach for extracting actionable insights from business data, there are several considerations to take into account:

1. Data quality and preprocessing: It is crucial to ensure that the business data used for analysis is of high quality, free from errors, inconsistencies, and biases.
2. Domain knowledge: It is essential to have a deep understanding of the specific business domain and its unique requirements. This knowledge will inform the design of the rule-based system and help identify relevant features and patterns in the data.
3. Scalability and computational resources: The hybrid approach may require significant computational resources, especially when working with large-scale textual datasets.

While the hybrid LLM-powered and rule-based approach may seem promising, it is essential to consider the potential drawbacks and limitations of this methodology. One of the primary concerns with relying on LLMs and large language models for data extraction is the ethical implications associated with using such powerful and complex AI systems.

Large language models, especially those trained on large-scale textual datasets, have raised concerns about perpetuating biases and misinformation present in the training data. The potential for these models to generate biased or inaccurate insights based on flawed linguistic characteristics is a significant risk that needs to be carefully addressed. Additionally, the complex nature of LLMs may introduce challenges in understanding and validating the decision-making process, especially when combined with rule-based systems.

Moreover, while LLMs have demonstrated impressive capabilities for natural language understanding and generation, there is a continuous need for extensive computational resources to train, fine-tune, and maintain these models. This reliance on substantial computational infrastructure may pose challenges for businesses with limited resources or computing capabilities, making the practical implementation of this approach a potential barrier.

## 3. Introduction to Business Insight Generation Data Pipeline

In the field of business intelligence, the journey from raw data to actionable insights involves a sophisticated pipeline that meticulously processes, analyzes, and synthesizes data. This pipeline is pivotal for organizations aiming to harness their data for strategic decision-making. While both rule-based and LLM approaches offer significant merits in data processing and analysis, the most effective business insight generation pipelines employ a hybrid strategy. This combines the precision and reliability of rule-based systems with the contextual understanding and linguistic flexibility of LLMs. Such a combination ensures a comprehensive analysis that is not only accurate but also richly informative and readily actionable.

### 3.1 Data Preprocessing

The foundation of any insightful analysis is high-quality data. In the data preprocessing stage, raw data is cleaned, normalized, and transformed to ensure consistency and relevance for subsequent analysis. Both rule-based methods and LLMs play crucial roles here. Rule-based approaches excel in systematically cleaning and structuring data according to predefined standards, while LLMs can offer additional support, particularly in interpreting and correcting unstructured textual data. The synergy of these methods ensures a robust preparation of data, laying the groundwork for insightful extraction.

**Importance:**

**Improves Data Quality**: Preprocessing cleans the data by fixing or removing inaccuracies and inconsistencies, significantly improving its quality.

**Ensures Consistency**: Normalization ensures that data from different sources or formats is brought to a common standard, facilitating accurate comparison and analysis.

**Enhances Model Performance**: Quality preprocessing directly impacts the performance of predictive models and analyses by providing them with reliable data.

**Challenges:**

**Variability of Data Sources**: Data can come from diverse sources with different formats and standards, making preprocessing complex.

**Missing Values**: Determining the best way to handle missing data—whether to impute, ignore, or remove it—can significantly affect the analysis outcomes.

**Scalability**: As datasets grow in size, preprocessing steps must scale accordingly, requiring efficient algorithms and processing power.

### 3.2 Data Preprocessor Structure

A data preprocessor is typically structured to sequentially apply a series of steps to clean and normalize the data. These steps might include:

**Data Cleaning**: Removing duplicates, correcting errors, and dealing with missing values.

**Data Integration**: Combining data from different sources into a cohesive dataset.

**Data Transformation**: Converting data into a format or scale suitable for analysis, such as normalizing ranges or encoding categorical variables.

**Data Reduction**: Reducing the dataset size through methods like principal component analysis (PCA) or feature selection to focus on the most informative aspects.

### 3.3 Rule-Based Approach

In a rule-based approach to preprocessing, specific rules and criteria are defined to handle different preprocessing tasks. For instance, rules could dictate that all missing values in a particular column should be replaced with the median value of that column, or that certain outlier values should be capped at a predefined threshold. This approach is highly structured and can be very efficient for datasets with well-understood characteristics and common preprocessing needs.

**Advantages:**

**Consistency and Control**: Offers consistent results and allows for fine-grained control over the preprocessing logic.

**Efficiency**: Can be highly efficient for datasets where the preprocessing needs are well understood and stable over time.

**Challenges:**

**Flexibility**: Adapting to new data sources or changes in the data can require manual updates to the preprocessing rules.

**Complexity**: Developing and maintaining a comprehensive set of preprocessing rules can become complex for large or diverse datasets.

### 3.4 LLM-Based Approach

An LLM-based approach to preprocessing involves leveraging language models to understand and manipulate data. This could involve using an LLM to infer missing values based on the context within the dataset or to identify and correct inconsistencies in textual data. While more experimental and less common than rule-based preprocessing, LLMs offer intriguing possibilities for handling complex and unstructured data.

**Advantages:**

**Adaptability**: Can adapt to new data patterns and inconsistencies without predefined rules.

**Handling Unstructured Data**: Particularly effective for preprocessing textual data, where LLMs can understand and correct nuances in language.

**Challenges:**

**Resource Intensity**: Requires significant computational resources, especially for large datasets.

**Predictability**: The outcomes of LLM-based preprocessing may be less predictable and harder to control than rule-based approaches.

**Conclusion**

Data preprocessing and normalization are critical steps that significantly impact the success of subsequent data analysis and modeling efforts. The choice between rule-based and LLM-based approaches depends on the specific characteristics of the data, the available resources, and the desired level of control over the preprocessing process. While rule-based approaches offer predictability and efficiency for structured datasets, LLM-based approaches provide flexibility and powerful capabilities for dealing with complex and unstructured data.

# 3.5 Experimental Approach: Data preprocessor built by LLM using input and output dataset examples

While full pre-processing of the input data stream might seem too resource intensive, one may consider using LLM's code generation capabilities. Building a data preprocessor in such a way involves designing a system that can analyze input dataset examples and understand the transformations needed to produce the desired output dataset. This process typically requires a combination of automated analysis and human oversight to ensure accuracy and relevance. Below, I'll outline a high-level approach to creating such a preprocessor, focusing on leveraging an LLM's capabilities to understand and apply data transformations based on examples.

**Step 1: Define the Input and Output Dataset Examples**

First, clearly define and document examples of your input and output datasets. These examples should illustrate the types of data preprocessing tasks you need to perform, such as handling missing values, normalizing data, transforming text, or categorizing information. The more comprehensive and varied your examples, the better the LLM can learn the desired transformations.

**Step 2: Design the Preprocessing Task Framework**

Develop a framework that outlines the types of preprocessing tasks your system should be able to handle. This framework could include:

- **Data Cleaning**: Identifying and correcting inaccuracies or inconsistencies.
- **Data Normalization**: Scaling numerical values or standardizing text formats.
- **Feature Engineering**: Deriving new data columns from existing ones based on specific logic.
- **Data Augmentation**: Generating synthetic data or additional features based on the input data.

**Step 3: Utilize the LLM for Task Identification**

With your examples and framework in place, use the LLM to identify the specific preprocessing tasks required for each example. This step involves querying the LLM with pairs of input and output examples and asking it to describe the transformations that occurred. For instance:

What preprocessing steps are needed to transform Dataset A (input) into Dataset B (output)? Identify and explain the data normalization techniques applied from Dataset A to B.

**Step 4: Generate Preprocessing Scripts or Commands**

Once the LLM has identified the necessary tasks, the next step is to generate the actual code or commands that perform these transformations. This can be achieved by querying the LLM with specific preprocessing tasks identified in the previous step and requesting code snippets in your language of choice (e.g., Python, SQL).

For example, ask the LLM to generate a Python function that applies the identified normalization technique to a given column of a pandas DataFrame.

**Step 5: Validate and Refine the Generated Code**

After generating the initial preprocessing scripts, it's crucial to validate their effectiveness on your datasets. This involves:

**Testing**: Run the generated code on your input datasets and compare the results with your output examples to ensure accuracy.
**Refinement**: If discrepancies are found, refine your queries to the LLM or adjust the generated code manually. This may involve providing the LLM with feedback on what was incorrect or asking for alternative solutions.

**Step 6: Automate and Iterate**

As you refine your preprocessing scripts, consider automating the process of querying the LLM and applying the generated transformations to new datasets. This could involve creating a pipeline that takes new data as input, uses the LLM to identify necessary preprocessing steps, generates the corresponding code, and applies it to the data.

**Step 7: Continuous Learning and Improvement**

Data preprocessing needs can evolve over time as new types of data are collected or analysis goals change. Continuously monitor the performance of your preprocessing system, and use new input-output examples to teach the LLM new transformations or refine existing ones. This ensures that your system remains effective and adaptable to changing requirements.

## Conclusion

Building a data preprocessor via an LLM based on input and output dataset examples represents a novel approach to automating data preprocessing tasks. While promising, this method requires careful design, testing, and refinement to ensure it meets the specific needs of your data analysis projects. By leveraging the power of LLMs to understand and generate code for data transformations, you can create a flexible and powerful tool to streamline your data preprocessing workflows.

## 4. Business Insights Extraction

Extracting meaningful business insights from processed data is a complex task that requires discernment and precision. This stage involves identifying patterns, anomalies, and trends that are significant to the business. Employing a hybrid approach, rule-based systems can efficiently sift through large datasets to pinpoint specific information based on established criteria, while LLMs can further analyze these findings to uncover deeper insights and context. This dual strategy allows for a thorough exploration of the data, ensuring that no valuable insight is overlooked.

**Examples of Business Insights**

In the domain of business intelligence, the insights gleaned from data analysis serve as critical inputs for strategic decision-making. The emergence of sophisticated analytics and machine learning technologies has empowered businesses to extract a wide array of valuable insights from their data repositories. In this section, we will explore various categories of business insights related to business metrics that are pivotal for organizations.

**General Anomalous Measurement Shifts**

Identifying general anomalous shifts in measurements across the entire dataset is crucial for early detection of issues or opportunities. These anomalies could signal a sudden change in consumer behavior, operational hiccups, or emerging market trends. By monitoring for unexpected deviations from historical patterns, businesses can swiftly respond to mitigate risks or capitalize on new developments.

**Specific Dimensions with Anomalous Measurement Shifts**

Drilling down from general anomalies, it's important to identify anomalous shifts within specific dimensions of the data. This could involve particular product lines, geographic regions, or customer segments exhibiting unusual patterns. Pinpointing these dimensions enables businesses to address the root causes of anomalies and tailor their strategies to specific aspects of their operations.

**Measurement Spikes**

Measurement spikes are sudden, sharp increases in specific metrics followed by a fast decrease, which could indicate both positive and negative developments. For instance, a spike in website traffic might result from a successful marketing campaign, while a surge in customer service complaints could highlight issues with a product or service. Recognizing and understanding the context of these spikes is essential for effective management and decision-making.

**All-Time High Measurement Values**

Achieving all-time high values in certain measurements, such as sales, user engagement, or production efficiency, is a clear indicator of business success. These milestones provide valuable insights into what strategies are working and serve as a benchmark for future performance. Celebrating these achievements can also boost morale and motivate teams to continue striving for excellence.

**Specific Dimensions with Top Values per Measurement**

Identifying which specific dimensions (e.g., product categories, regions, sales channels) are performing the best in terms of specific measurements can inform resource allocation and strategic focus. For example, if certain products are consistently top sellers, a business might decide to expand those lines or explore similar market opportunities.

**Specific Dimensions Comparison per Business Performance**

Comparing specific dimensions in relation to overall business performance allows for a nuanced understanding of how different areas of the business contribute to success. This can involve comparing sales performance across regions, customer satisfaction by product line, or marketing ROI by channel. Such comparisons not only highlight areas of strength but also reveal potential improvement opportunities.

Incorporating these types of insights into business strategies enables organizations to navigate complex markets with greater agility and precision. By leveraging data analytics and machine learning, businesses can transform raw data into actionable intelligence, driving growth and competitive advantage in today's data-driven world.

## 5. An Overview of Various Approaches to Extracting Business Insights

The process of business insights extraction from the preprocessed structured data can be approached through rule-based systems or LLMs, each offering distinct advantages and facing unique challenges. Rule-based approaches are highly effective in structured data environments, offering high precision, resource efficiency, deterministic outcomes, ease of interpretability, and customizability for domain-specific needs. However, they may struggle with scalability, flexibility, complexity in rule creation, overlooking nuanced patterns, and require significant maintenance overhead as data environments evolve.

On the other hand, LLMs provide adaptability to new data patterns, excel in handling unstructured data, and can generate rich, nuanced insights. They also reduce the maintenance overhead associated with rule updates. Yet, LLMs face challenges such as high resource intensity, interpretability issues, less precision in highly structured data, significant training data requirements, risk of bias, and difficulties in performing precise mathematical operations. The juxtaposition of these approaches highlights a landscape where the integration of rule-based systems and LLMs can offer a comprehensive solution, leveraging the precision and reliability of rule-based methods with the flexibility and depth of LLM-generated insights. This hybrid approach aims to balance the strengths and mitigate the challenges inherent in each method, providing a robust framework for business insights extraction that is adaptable, scalable, and aligned with organizational goals.

## 5.1 Rule-Based Approaches to Business Insights Extraction

From an engineering perspective, rule-based approaches to business insights extraction operate on a framework of predefined logic and criteria to process and analyze data. This methodology is deeply rooted in the principles of traditional programming and data processing, where every operation is explicitly defined by the developers or data scientists which requires a solid foundation in programming, data science, and domain-specific knowledge. The effectiveness of these systems depends on the precision with which rules are defined and applied, as well as the system's ability to process and analyze data efficiently and accurately.

**Advantages**

**High Precision in Structured Data**: Rule-based systems excel in environments with structured data, where precise conditions and thresholds can be defined for insight extraction, ensuring high accuracy in identifying specific patterns or anomalies.

**Resource Efficiency**: Compared to LLM approaches, rule-based systems generally require fewer computational resources for processing structured data, making them more cost-effective for certain types of analysis.

**Deterministic Outcomes**: The deterministic nature of rule-based systems guarantees consistent results, which is crucial for repeatable analysis and tracking changes over time within structured datasets.

**Ease of Interpretability**: Insights generated through rule-based methods are easier to trace back to their originating logic, offering clear interpretability and the ability to easily validate findings against business logic.

**Customizability for Domain-Specific Needs**: Rule-based systems can be finely tuned to specific business contexts and domains, allowing for tailored insight extraction that aligns closely with organizational goals and data characteristics.

**Challenges**

**Scalability Limitations**: As datasets grow in size and complexity, maintaining and updating the rules can become increasingly challenging, limiting the scalability of rule-based systems.

**Flexibility and Adaptability**: Rule-based systems can struggle to adapt to new patterns or changes in data structure without manual intervention, potentially missing emerging insights not covered by existing rules.

**Complexity in Rule Creation**: Developing comprehensive and effective rules requires deep domain expertise and understanding of the data, which can be resource-intensive and time-consuming.

**Overlooked Nuances**: Rule-based systems might overlook subtler, complex patterns in data that do not fit neatly into predefined criteria, potentially missing valuable insights.

**Maintenance Overhead**: As business contexts and data environments evolve, rule-based systems require continuous review and updates to rules, creating significant maintenance overhead.

## 5.2 LLM Approaches to Business Insights Extraction

LLM approaches to business insights extraction represent a shift towards leveraging advanced artificial intelligence to analyze and interpret data. Unlike rule-based systems, LLMs rely on pre-trained models that understand and generate natural language, allowing for a more nuanced and context-aware analysis of data.

**Advantages**

**Adaptability to New Patterns**: LLMs can quickly adapt to new data patterns and changes, providing the flexibility to generate insights from evolving datasets without the need for manual rule adjustments.

**Handling of Unstructured Data**: Beyond structured data, LLMs excel in extracting insights from unstructured data, offering a broader scope of analysis that includes textual analysis and sentiment extraction.

**Richness of Insights**: LLMs can generate more nuanced and contextually rich insights, capturing complex relationships in the data that might be missed by rule-based systems.

**Reduced Maintenance Overhead**: Once trained, LLMs can continue to provide insights without the same level of ongoing maintenance and rule updates required by rule-based systems.

**Challenges**

**Resource Intensity**: LLMs typically require significant computational resources for training and inference, especially when processing large datasets, which can be costly.

**Interpretability Issues**: Insights generated by LLMs may not always offer clear interpretability, making it challenging to understand the rationale behind certain analyses.

**Precision in Structured Data**: For highly structured datasets, LLMs may not always match the precision of rule-based systems, especially in scenarios requiring strict adherence to predefined conditions.

**Training Data Requirements**: LLMs require large amounts of training data to perform optimally, which can be a limitation in data-scarce environments or when dealing with highly specific business contexts.

**Risk of Bias**: If not carefully managed, LLMs can perpetuate or amplify biases present in their training data, leading to skewed insights and potential ethical concerns.

**Mathematical Operations**: LLMs might struggle with performing precise mathematical operations or extracting insights based on complex numerical analysis, a task for which rule-based systems with explicitly defined logic can be a lot more suited and reliable.

## 6. Natural Language Narrative Generation

Transforming data-driven insights into natural language narratives makes the analysis accessible and understandable to non-technical decision-makers or in other words democratizes the data. LLMs are particularly adept at this task, utilizing their advanced natural language generation capabilities to articulate complex insights in clear, concise language. However, incorporating rule-based logic can enhance this process by structuring the narratives around key business metrics and objectives, ensuring that the generated text aligns with specific analytical goals. The result is a set of narratives that not only convey the insights but do so in a manner that is directly relevant to the business's strategic interests.

## 6.1 Rule-Based Approaches to Natural Language Narrative Generation

From an engineering perspective, rule-based approaches to natural language narrative generation involve a structured process where explicit rules dictate how data insights are translated into text. This method leverages a combination of data processing techniques and linguistic rules to produce narratives that are both informative and aligned with specific analytical goals.

**Advantages**

**Precision and Relevance**: Rule-based systems can generate narratives that precisely match specific business metrics and objectives, ensuring that every piece of generated text is highly relevant and aligned with predefined analytical goals.

**Efficiency in Structured Environments**: For structured data insights, rule-based narrative generation can be more resource-efficient, as it operates within well-defined parameters without the need for extensive computational power.

**Consistency Across Narratives**: By adhering to a set of predefined rules, this approach guarantees a high level of consistency in the narrative output, crucial for maintaining a uniform tone and style across all generated reports.

**Customization to Business Needs**: Rule-based systems allow for extensive customization, enabling the engineering of narratives that cater to the unique needs and preferences of different business stakeholders.

**Deterministic Outcomes**: The deterministic nature ensures that given the same set of insights, the narrative output will be consistent, providing a reliable basis for decision-making and reporting.

**Challenges**

**Limited Scalability and Flexibility**: As datasets grow and evolve, updating and maintaining the rule set for narrative generation can become increasingly complex, limiting the system's scalability and flexibility.
**Complexity in Rule Development**: Creating and refining the rules for narrative generation requires a deep understanding of both the domain and natural language structures, making the process resource-intensive.

**Risk of Overlooking Nuances**: This approach may miss subtleties in the data or fail to capture the full context of insights, leading to narratives that lack depth or fail to engage the audience.

**Maintenance Overhead**: Continuous monitoring and updating of the rule set to align with changing business conditions and data environments create a significant maintenance burden.

**Rigid Narrative Structures**: Rule-based narrative generation might result in texts that are structurally rigid and lack the fluidity or creativity that can engage readers more effectively.

## 6.2 LLM Approaches to Natural Language Narrative Generation

Large Language Models (LLMs) represent a significant advancement in natural language processing (NLP) and generation (NLG), offering a powerful tool for transforming data-driven insights into natural language narratives. From an engineering perspective, deploying LLMs for narrative generation involves several key steps, leveraging the models' ability to understand context, generate coherent text, and adapt to new information.

**Advantages**

**Adaptability to Evolving Data**: LLMs can adapt to new patterns and insights from evolving datasets, generating narratives that reflect the most current context without manual adjustments.

**Richness and Nuance in Narratives**: Leveraging advanced NLP capabilities, LLMs can produce narratives that capture the subtleties and complexities of the insights, providing depth and engaging the audience more effectively.

**Efficiency at Scale**: LLMs can handle large volumes of data and generate narratives at scale, benefiting from their ability to process and synthesize information quickly.

**Reduced Maintenance**: Once an LLM is trained or fine-tuned, it requires less ongoing maintenance compared to rule-based systems, as it can automatically adapt to new information patterns.

**Creative and Engaging Text Generation**: LLMs have the potential to generate narratives that are not only informative but also engaging and creative, enhancing the readability of insight reports.

**Challenges**

**High Resource Requirements**: Training and running LLMs, especially for narrative generation from complex insights, require substantial computational resources and associated costs.

**Interpretability and Transparency Issues**: Understanding how an LLM constructs narratives from data insights can be challenging, raising questions about the process's transparency. Precision in Structured Data Narratives: In scenarios requiring strict adherence to business metrics, LLMs might not match the precision offered by rule-based systems, potentially affecting the accuracy of the narratives.

**Quality and Relevance of Training Data**: The effectiveness of LLMs in generating relevant and accurate narratives heavily depends on the quality and domain relevance of the training data.

**Risk of Bias and Inaccuracy**: There is a potential for LLMs to perpetuate biases present in their training data or generate inaccuracies due to overgeneralization, affecting the credibility of the narratives.

# 7. Architectures of Hybrid Data Processing Pipelines for Business Insight Generation

To address the challenges arising at various stages of business insight generation, innovative data processing pipeline architectures have been developed, combining the strengths of rule-based systems with the advanced capabilities of LLMs. These hybrid approaches offer a sophisticated solution to the complexities of business intelligence, balancing precision with the nuance of natural language understanding. Various architectures of that kind [6] point out using structured information alongside the LLMs in graph related contexts [8] suggest a multi-stage LLM based algorithm with results reranking, more approaches mentioned in the References section, all having one thing in common – LLM is not treated as a static black box that takes in prompts and outputs text, but rather one of the elements of complex processing pipelines with rule base and algorithmic input and output curation.

**The first architecture (LLM-Based Insight Generation from Chunked Data)** introduces a method where unprocessed data is segmented and fed through an LLM alongside expertly crafted prompts. This "chunking" strategy is designed to circumvent the token input limitations of LLMs, enabling the analysis of extensive datasets while focusing on generating relevant insights. It exemplifies the adaptability and scalability of combining traditional data processing techniques with the power of LLMs.

**Our second architecture (Sequential Data Processing and Insight Generation)** outlines a sequential approach that begins with meticulous data preprocessing and moves towards extracting specific data fragments for analysis. These fragments are then enriched with expert prompts that guide the LLM in producing atomic insights, which are synthesized into a final report. This process highlights the precision of targeted analysis and the LLM's ability to generate insights from carefully curated data fragments, illustrating the benefits of a methodical, step-by-step approach to data analysis.

**The third architecture (Hybrid Rule-Based and LLM Insight Generation)** presents a compelling hybrid model, utilizing a rules-based engine for the initial generation of atomic business insights, followed by the summarization capabilities of an LLM to craft these insights into a cohesive, well-articulated report. This architecture leverages the accuracy and reliability of rule-based analysis with the advanced natural language generation skills of LLMs, offering a balanced solution that combines the best of both worlds.

These hybrid architectures underscore the significant advantages of integrating rule-based systems with LLMs over relying on purely rule-based or LLM methods alone. By doing so, organizations can achieve a higher level of precision and detail in their analysis while also benefiting from the contextual understanding and linguistic sophistication of LLMs. This synergy not only enhances the quality of insights generated but also ensures that these insights are both actionable and accessible to decision-makers. As we explore these architectures further, it becomes evident that the future of business intelligence and data analysis will increasingly rely on such hybrid approaches, which offer a comprehensive solution to the complex challenge of transforming data into strategic business insights.

## 7.1 LLM-Based Insight Generation from Chunked Data

This architecture involves directing unprocessed, arbitrary data through a Large Language Model (LLM) to generate natural language insights, utilizing expertly crafted prompts to guide the analysis. Given the LLM's token input limitations — where the amount of data that can be processed in a single prompt is restricted — the proposed solution includes a "chunking" strategy. This approach entails dividing the large dataset into smaller, manageable pieces or "chunks" that are processed sequentially or in parallel by the LLM, with each chunk accompanied by a tailored prompt designed to extract specific insights or information.

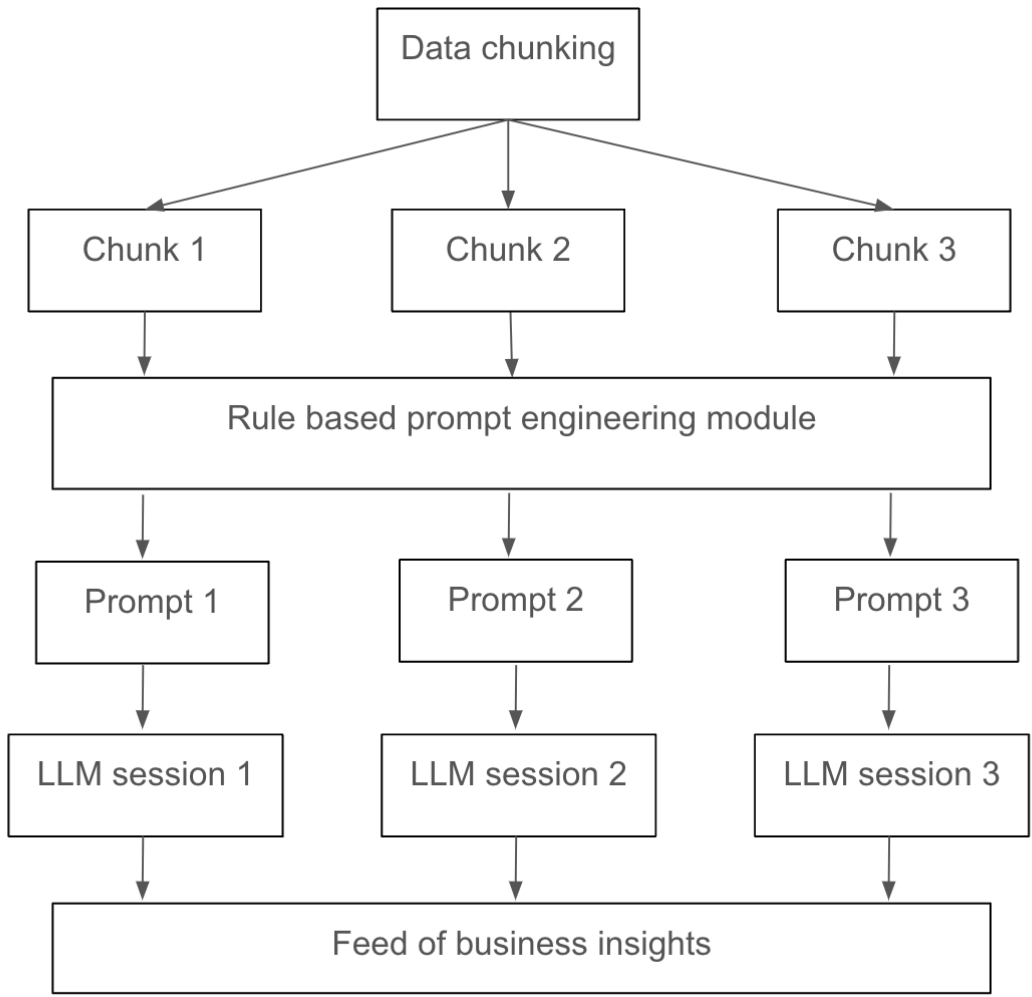

*Fig 1. LLM-Based Insight Generation from Chunked Data*

**Data Preparation**: The unprocessed dataset is divided into smaller chunks. This division can be based on logical segments of the data, such as temporal splits (e.g., monthly data), categorical divisions (e.g., by product line or region), or simply by breaking the dataset into parts that fit within the LLM's token limits.

**Prompt Engineering**: For each chunk, an expertly designed prompt is created. These prompts are crafted to direct the LLM's focus towards extracting relevant insights from the specific portion of data it receives, taking into account the context and goals of the analysis.

**Sequential or Parallel Processing**: Depending on the infrastructure and urgency, the data chunks can be processed either sequentially or in parallel. Parallel processing significantly speeds up the analysis but requires more computational resources.

**Insight Generation**: The LLM generates insights for each chunk, which are then compiled and synthesized into a comprehensive analysis.

**Advantages**

**Scalability**: By breaking down the dataset, this approach can handle large volumes of data that exceed the LLM's token limits, making it scalable to various data sizes.

**Focused Analysis**: Chunking allows for tailored prompts that can direct the LLM to generate more relevant and focused insights for each segment of the data.

**Parallel Processing Capability**: The ability to process data chunks in parallel can significantly reduce the time required for insight generation, making this approach efficient for time-sensitive analyses.

**Challenges**

**Integration of Insights**: Combining insights from different chunks into a coherent and comprehensive analysis can be challenging, especially if the chunks are processed independently without consideration of the overall context.

**Chunking Strategy**: Determining the optimal way to divide the data into chunks requires careful consideration to ensure that important patterns or trends are not overlooked. Poor chunking strategies can lead to fragmented insights that miss the bigger picture.

**Resource Intensity**: While parallel processing offers speed advantages, it also demands significant computational resources, which may not be feasible for all organizations or scenarios.

**Prompt Engineering Complexity**: Crafting effective prompts for each chunk requires a deep understanding of both the data and the analysis objectives, making prompt engineering a potentially complex and time-consuming task.

**Conclusion**

The chunking approach to utilizing LLMs for insight generation from large, unprocessed datasets offers a scalable and flexible solution to the challenge of LLM token limitations. While it presents several advantages in terms of focused analysis and processing efficiency, it also introduces complexities related to insight integration, chunking strategy, and the need for expert prompt engineering. Addressing these challenges requires careful planning and potentially innovative solutions to ensure that the final insights are both comprehensive and actionable.

## 7.2 Sequential Data Processing and Insight Generation

This architecture outlines a structured approach to generating business insights from data, starting with preprocessing and culminating in a comprehensive insight report. The process emphasizes the extraction of specific data fragments, enrichment with expertly crafted prompts, and the use of a Large Language Model (LLM) to produce atomic insights, which are then synthesized into a final report. This methodology aims to ensure that the insights are both precise and relevant to the specific questions at hand.

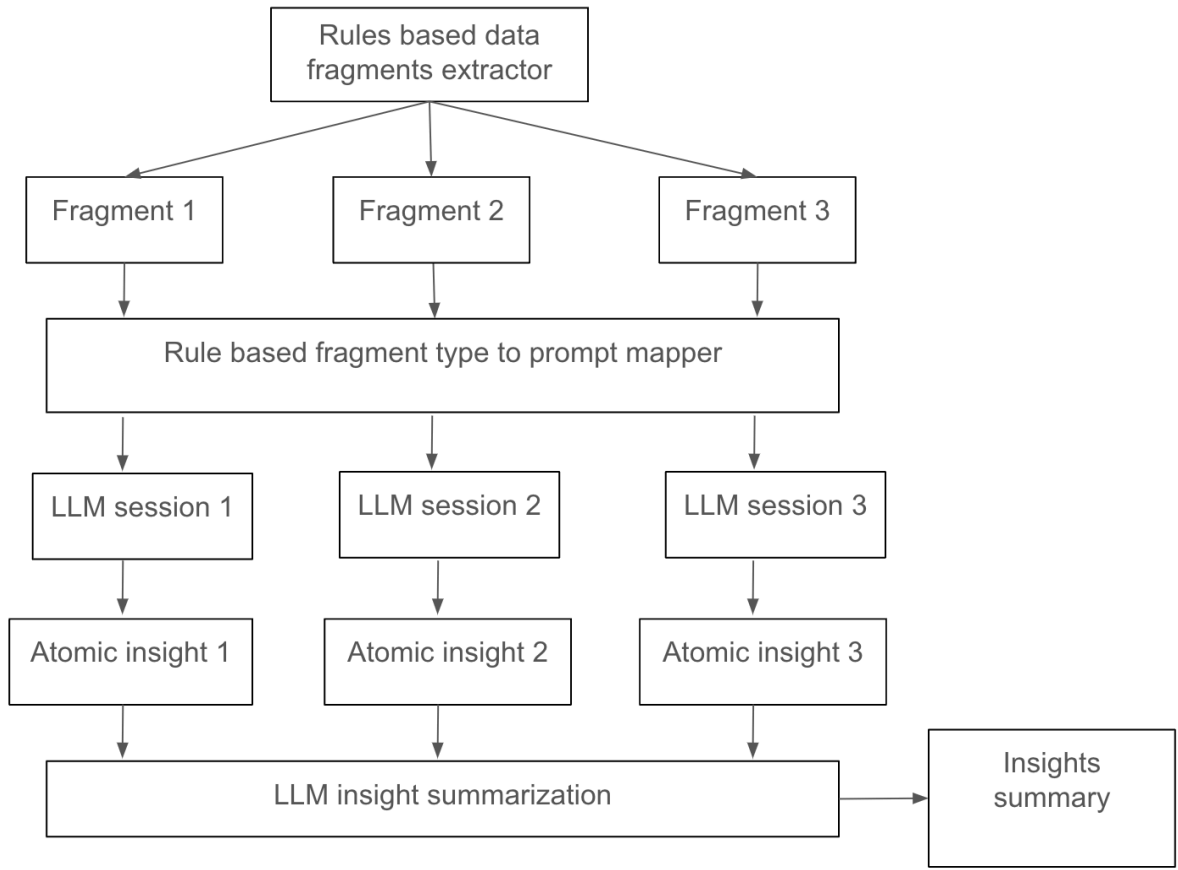

*Fig 2: Sequential Data Processing and Insight Generation*

**Step 1: Data Preprocessing**

The initial step involves cleaning, normalizing, and preparing the data for analysis. This includes removing duplicates, handling missing values, normalizing data ranges, and possibly transforming the data to ensure consistency and accuracy. Effective preprocessing lays the groundwork for more accurate and meaningful insights by ensuring the data is in a suitable format for analysis.

**Step 2: Extraction of Specific Fragments**

Following preprocessing, the next step is to extract specific fragments of the data that are relevant to the questions being asked. This targeted approach ensures that the analysis is focused and efficient, dealing only with data that can potentially yield useful insights. The selection of fragments can be based on certain criteria or conditions that align with the business questions of interest.

**Step 3: Enrichment with Expert Prompts**

The extracted data fragments are then enriched with prompts created by domain experts. These prompts are designed to guide the LLM in analyzing the data fragments, directing it to focus on generating insights that are relevant to the specific business questions. The prompts act as a bridge between the raw data and the LLM's capability to generate insightful analysis, ensuring that the model's output is aligned with the business objectives.

**Step 4: Generation of Atomic Business Insights**

With the prompts in place, the LLM is employed to analyze each enriched data fragment and generate atomic business insights. These insights are "atomic" in the sense that each one addresses a specific aspect or question related to the broader topic of interest. The use of an LLM facilitates the generation of nuanced and contextually relevant insights that might not be readily apparent through traditional analysis methods.

**Step 5: Summarization into the Final Insight Report**

The final step involves synthesizing the atomic insights into a comprehensive insight report. This synthesis requires careful consideration of how each atomic insight fits into the overall picture, ensuring that the final report provides a coherent and comprehensive analysis of the data in relation to the business questions. The summarization process may also involve prioritizing certain insights, drawing connections between them, and presenting them in a format that is accessible and actionable for decision-makers.

**Advantages**

**Targeted Analysis**: By focusing on specific data fragments and questions, this approach ensures that the analysis is highly relevant and efficient.

**Expert Guidance**: The use of expert-crafted prompts ensures that the LLM's analysis is closely aligned with the business objectives, enhancing the relevance and usefulness of the insights.

**Depth of Insight**: Leveraging an LLM to generate insights allows for a level of depth and nuance that is difficult to achieve with traditional analysis methods.

**Challenges**

**Complexity of Prompt Engineering**: Crafting effective prompts requires a deep understanding of both the domain and the questions at hand, which can be challenging and time-consuming.

**Integration of Insights**: Synthesizing atomic insights into a cohesive report requires a clear understanding of how each insight contributes to the overall analysis, which can be complex, especially for large datasets or multifaceted questions.

**Data Fragmentation**: Ensuring that the extraction of specific data fragments does not overlook important context or connections between different parts of the data can be challenging.

**Conclusion**

Architecture 2 offers a structured and targeted approach to generating business insights, leveraging the strengths of LLMs guided by expert input. While it presents several advantages in terms of producing deep and relevant insights, it also poses challenges related to the complexity of prompt engineering and the integration of insights. Addressing these challenges effectively is crucial for maximizing the value of the final insight report.

## 7.3 Hybrid Rule-Based and LLM Insight Generation

This architecture proposes a hybrid approach combining a rules-based engine for generating atomic business insights with the use of a Large Language Model (LLM) for summarizing these insights into a coherent, well-written final report. This methodology leverages the precision and reliability of rule-based systems for initial insight generation, while capitalizing on the strengths of LLMs in natural language understanding and generation for the creation of the final report.

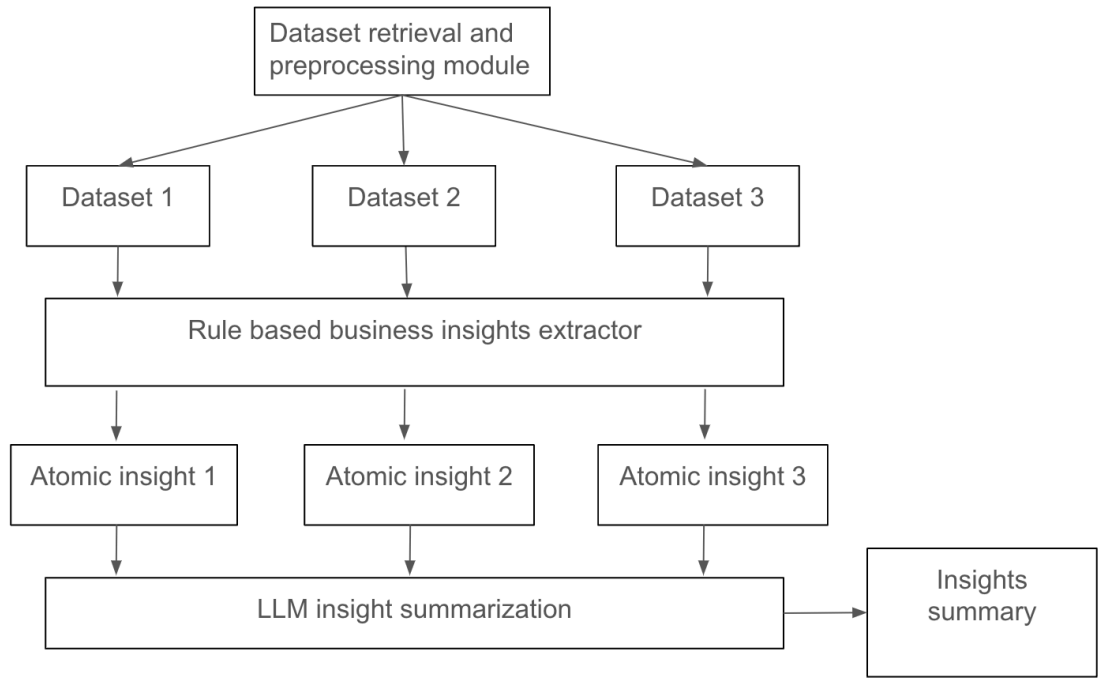

*Fig 3. Hybrid Rule-Based and LLM Insight Generation*

**Step 1: Data Preprocessing**

The process begins with thorough data preprocessing to ensure the quality and consistency of the dataset. This includes cleaning the data (removing duplicates, handling missing values), normalizing (standardizing formats and scales), and potentially transforming data (encoding categorical variables, generating new features) to prepare it for analysis. Effective preprocessing is critical to ensure that the insights generated in later stages are based on reliable and accurate data.

**Step 2: Rules-Based Engine for Atomic Insight Generation**

With the data prepared, a rules-based engine analyzes the dataset to generate atomic business insights. This engine operates on predefined logic and criteria to identify patterns, anomalies, trends, or other relevant findings within the data. The rules are crafted based on domain knowledge and specific business objectives, ensuring that the insights are directly applicable and valuable to the business. This approach benefits from the high level of control and transparency, allowing for consistent and interpretable insight generation.

**Step 3: LLM as a Summarizer for Final Report**

The atomic insights generated by the rules-based engine are then passed to an LLM, which acts as a summarizer. The LLM leverages its advanced natural language generation capabilities to synthesize the atomic insights into a well-structured, coherent final report. This report not only summarizes the insights but also contextualizes them, highlighting their significance and potential implications for the business. The LLM's ability to understand and generate natural language ensures that the final report is accessible and engaging for its intended audience.

**Advantages**

**Precision and Reliability**: The use of a rules-based engine for initial insight generation ensures that the insights are accurate, consistent, and based on well-defined criteria.

**High-Quality Reporting**: Leveraging an LLM for report generation capitalizes on the model's strengths in summarization and natural language production, resulting in a final report that is both informative and well-written.

**Efficient Process**: This hybrid approach allows for an efficient division of labor between the rules-based engine and the LLM, with each component focusing on what it does best.

**Challenges**

**Complexity in Integration**: Ensuring seamless integration between the rules-based insight generation and the LLM summarization can be complex, requiring careful design to ensure that the insights are accurately and effectively communicated to the LLM.

**Rules Maintenance**: The rules-based engine requires ongoing maintenance and updating to remain aligned with evolving business objectives and data characteristics.

**Summarization Accuracy**: While LLMs are generally good at summarization, ensuring that the final report accurately reflects the nuances and importance of the atomic insights requires careful prompt engineering and possibly manual review.

**Conclusion**

Architecture 3 presents a strategic approach to business insight generation, combining the strengths of rules-based engines and LLMs to produce actionable and well-communicated insights. This hybrid model offers a balance between the precision of rule-based analysis and the linguistic prowess of LLMs, making it a powerful tool for businesses looking to derive meaningful insights from their data. Addressing the integration and maintenance challenges inherent in this approach is crucial for its success, ensuring that businesses can leverage their data effectively to inform strategic decision-making.

# 8. Benchmarking

We benchmarked the efficiency of data analysis and specifically extraction of important business events from the business related time series datasets in the form of readable insights using 3 basic approaches for each case: purely rule based, LLM and some form of hybrid approach. The data used for the benchmarking was collected from 30 corporate Google Analytics 4 and Google Ads accounts via APIs for the time frame of approximately two years. The LLM used for the research was GPT-4 accessed via its native API.

### 8.1 Precision of mathematical operations

LLMs can perform some mathematical calculations and do certain logical reasoning, although the precision of those operations is not absolute due to a number of factors.

- Calculations is not a fundamental feature of LLMs but rather a side effect with no guaranteed quality

- Certain metrics have specific calculation requirements, i.e. can't be treated in a generic way, e.g. the average value for the cost-per-click type of metrics can't be calculated using the standard average formula but needs the weighted average schema to be applied

To minimize the number of LLM induced errors a rule based preprocessing algorithm is added to the analytical pipeline which pre-calculates total and average values for each business metric specifically for the required time periods. It increases the prompt size therefore reducing the amount of data that can be sent into the LLM which is a typical trade off, i.e. recall traded off for precision.

| Processing pipeline type | Processing efficiency |
| --- | --- |
| Rule based | 100% |
| LLM | 63% |
| Hybrid (Rule based precalculation + LLM analysis) | 87% |

### 8.2 Number of proper name hallucinations

LLMs can sometimes "hallucinate" names or facts, producing outputs that seem plausible but are incorrect or fictional. This is particularly challenging when dealing with data that includes proper names, where accuracy is crucial.

- Rule-based: This approach relies solely on pre-defined rules and does not generate new content, thus minimizing the risk of hallucinating names. It would typically result in a very low or even zero count of proper name hallucinations.

- LLM: Given its generative nature, LLM is more prone to hallucinate names compared to rule-based systems. While it's sophisticated enough to generate highly plausible content, distinguishing between real and generated names without external validation can be challenging.

- Hybrid (Name hashing + LLM analysis + Hash decoding): This approach uses a combination of name hashing to anonymize real names before processing with LLM, followed by hash decoding to restore names in the final output. This can significantly reduce the number of proper name hallucinations since the model isn't directly generating or manipulating the proper names.

| Processing pipeline type | Number of errors |
|---|---|
| Rule based | 0% |
| LLM | 12% |
| Hybrid (Name hashing + LLM analysis + Hash decoding) | 3% |

## 8.3 Recall of the important business insights

Recall, in this context, refers to the system's ability to extract and present all relevant business insights from the data, a crucial factor for comprehensive analysis.

- Rule-based: This method might not capture all nuances or connections in the data due to its reliance on predefined patterns and thresholds, potentially leading to lower recall.

- LLMs are adept at identifying patterns and insights in large datasets, sometimes allowing for non-trivial pattern recognition and out-of-the-box reasoning, potentially discovering business insights that can't be accessed by rule-based systems. However, the quality of the insights can vary based on the input data and the model's current knowledge.

- Hybrid (Source specific data chunking + LLM analysis + LLM summarization of the processed chunks): By chunking the data and using LLM for both detailed analysis and summarization, this approach aims to combine the best of both worlds. It potentially offers high recall by leveraging LLM's pattern recognition and generative capabilities while focusing the analysis on manageable portions of the dataset.

| Processing pipeline type | Processing efficiency |
| --- | --- |
| Rule based | 71% |
| LLM | 67% |
| Hybrid (Source specific data chunking + LLM analysis + LLM summarization of the processed chunks) | 82% |

## 8.4 Overall user satisfaction on weekly/monthly reports

User satisfaction can be influenced by factors like the accuracy of the information, the relevance and comprehensiveness of the insights provided, and the readability of the reports. This metric is measured as the ratio of likes to dislikes (the bigger the number, the higher the overall user satisfaction).

- Rule-based: While highly accurate, rule-based reports might lack the narrative quality and comprehensive insights that come from more sophisticated analysis, possibly leading to lower user satisfaction if the reports seem too dry or limited.

- LLM can generate more engaging and detailed reports with a narrative structure, potentially leading to higher user satisfaction. However, the accuracy and relevance of the insights depend heavily on the input data and the model's training.

- Hybrid (Rule based + LLM analysis): This approach aims to combine the reliability of rule-based preprocessing with the narrative and analytical strengths of LLM, more specifically mixing the insight generation capabilities of the rule based module and data analysis/text summarization capabilities of the LLM.

| Processing pipeline type | Likes-to-dislikes ratio |
| --- | --- |
| Rule based | 1.79 |
| LLM | 3.82 |
| Hybrid | 4.60 |

## 9. Conclusion

In conclusion, the exploration of hybrid LLM-powered and rule-based systems for business insights generation from structured data reveals a compelling pathway towards more sophisticated, accurate, and actionable business intelligence. This paper has demonstrated that while each approach has its inherent strengths and weaknesses, their integration can harness the precision and reliability of rule-based systems alongside the dynamic and context-sensitive capabilities of LLMs. This synergy not only enhances the extraction and analysis of complex business data but also addresses some of the critical challenges faced by each method when applied independently.

The hybrid model proposes a balanced approach, enabling better handling of diverse and complex datasets while maintaining transparency and reducing the potential biases associated with AI-driven systems. Future research could focus on refining these hybrid models, optimizing computational resources, and expanding the types of business data that can be effectively analyzed.

As businesses continue to navigate increasingly data-driven landscapes, the role of advanced hybrid analytical systems will undoubtedly expand, becoming integral to strategic decision-making and operational efficiencies. The insights from this study provide a foundation upon which organizations can build more resilient and insightful data analysis practices, driving growth and innovation in an ever-evolving business environment

# References

1. Xiaoyu Liu, Hiba Alsghaier, Ling Tong, Amna Ataullah, Susan McRoy. Visualizing the Interpretation of a Criteria-Driven System That Automatically Evaluates the Quality of Health News: Exploratory Study of 2 Approaches (2022). https://scite.ai/reports/10.2196/37751
2. Konstantina Christakopoulou, Madeleine Traverse, Trevor Potter, Emma Marriott, Daniel Li, Chris Haulk, Ed H. Chi, Minmin Chen. Deconfounding User Satisfaction Estimation from Response Rate Bias (2020). https://scite.ai/reports/10.1145/3383313.3412208
3. A Benchmark To Understand The Role Of Knowledge Graphs On Large Language Model's Accuracy For Question Answering On Enterprise Sql Databases. https://arxiv.org/pdf/2311.07509.pdf
4. SQL-to-Text Generation with Graph-to-Sequence Model. https://arxiv.org/pdf/1809.05255v2.pdf
5. LLMs in HCI Data Work: Bridging the Gap Between Information Retrieval and Responsible Research Practices. https://arxiv.org/pdf/2403.18173.pdf
6. Can LLMs Effectively Leverage Graph Structural Information through Prompts, and Why? https://arxiv.org/pdf/2309.16595.pdf
7. A Hybrid Approach To Aspect Based Sentiment Analysis Using Transfer Learning. https://arxiv.org/pdf/2403.17254.pdf
8. Enhancing Legal Document Retrieval: A Multi-Phase Approach with Large Language Models. https://arxiv.org/pdf/2403.18093.pdf
9. ChainLM: Empowering Large Language Models with Improved Chain-of-Thought Prompting. https://arxiv.org/pdf/2403.14312.pdf
10. Methods and systems of facilitating provisioning contexts for business situations using a semantic graph.
    https://patents.google.com/patent/US20230410016A1
11. Generating actionable insight information from data sets using an artificial intelligence-based natural language interface.
    https://patents.google.com/patent/US20230281228A1